\documentclass[10pt,twocolumn,letterpaper]{article}

\usepackage{cvpr}              

\newcommand{\impro}[1]{{\hspace{0.05cm}{\color[HTML]{32CB00}\textbf{(+#1)}}}}

\definecolor{cvprblue}{rgb}{0.21,0.49,0.74}
\usepackage[pagebackref,breaklinks,colorlinks,allcolors=cvprblue]{hyperref}

\def\paperID{4329} 
\def\confName{CVPR}
\def\confYear{2026}

\usepackage{amssymb}
\usepackage{multicol}
\usepackage{multirow}
\usepackage{pifont}

\newcommand{\xmark}{\ding{55}}
\usepackage{colortbl}

\title{T2SGrid: Temporal-to-Spatial Gridification for Video Temporal Grounding}

\author{
Chaohong Guo$^{1}$
\quad  Yihan He$^{1}$
\quad Yongwei Nie$^{1}$\\ 
\quad Fei Ma$^{2}$
\quad Xuemiao Xu$^{1}$
\quad Chengjiang Long$^{3}$ \\ \\ 
\small $^{1}$South China University of Technology.
\\ \small $^{2}$Guangdong Laboratory of Artificial Intelligence and Digital Economy (SZ).
\\ \small $^{3}$Bytedance Inc. \\
\small \texttt{cguo5104@gmail.com}
}

\begin{document}
\maketitle
\begin{abstract}

Video Temporal Grounding (VTG) aims to localize the video segment that corresponds to a natural language query, which requires a comprehensive understanding of complex temporal dynamics. Existing Vision-LMMs typically perceive temporal dynamics via positional encoding, text-based timestamps, or visual frame numbering. However, these approaches exhibit notable limitations: assigning each frame a text-based timestamp token introduces additional computational overhead and leads to sparsity in visual attention, positional encoding struggles to capture absolute temporal information, and visual frame numbering often compromises spatial detail. 
To address these issues, we propose Temporal to Spatial Gridification (T2SGrid), a novel framework that reformulates video temporal understanding as a spatial understanding task. The core idea of T2SGrid is to process video content in clips rather than individual frames. we employ a overlapping sliding windows mechanism to segment the video into temporal clips. Within each window, frames are arranged chronologically in a row-major order into a composite grid image, effectively transforming temporal sequences into structured 2D layouts. The gridification not only encodes temporal information but also enhances local attention within each grid. Furthermore, T2SGrid enables the use of composite text timestamps to establish global temporal awareness. 
Experiments on standard VTG benchmarks demonstrate that T2SGrid achieves superior performance. Code is available at \url{https://github.com/chaohongguo/T2SGrid}.

\end{abstract}    
\section{Introduction}
\label{sec:intro}

\begin{figure}[t]
  \centering
   \includegraphics[width=\linewidth]{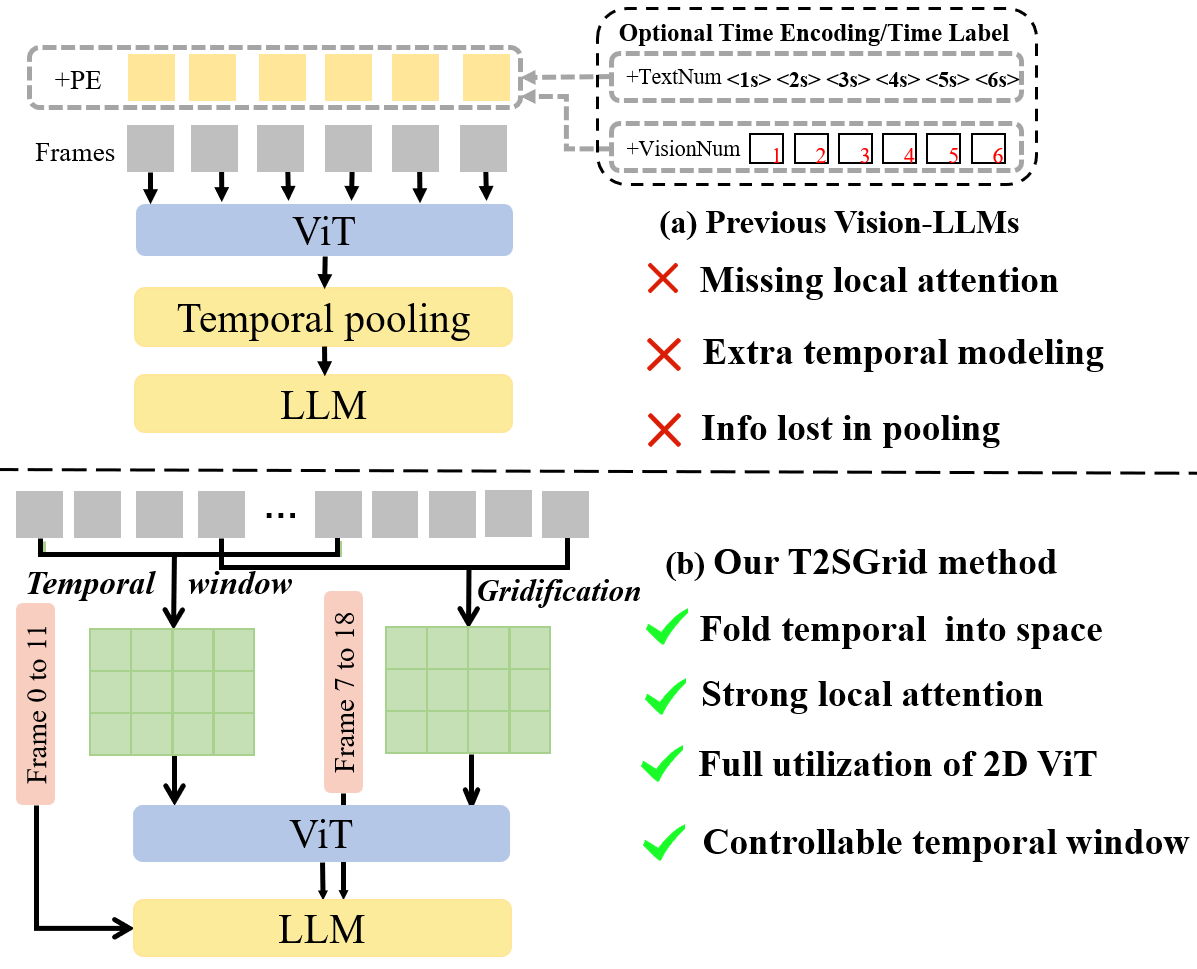}

   \caption{\textbf{Comparison between T2SGrid and previous Vision-LLMs.} (a) Traditional methods process frames sequentially or apply temporal pooling to capture information at multiple scales. However, sequential processing and pooling can cause information loss, obscure local temporal details within a window, and often require additional time encoding/label for temporal modeling.  (b) Our T2SGrid method folds multiple frames within a temporal window into the spatial dimension via gridification, allowing direct processing a temporal window by the standard 2D ViT. This leverages the model's strong spatial reasoning capability for temporal understanding.}
   \label{fig:intro}
\end{figure}

Video content has become a ubiquitous medium for information dissemination, yet efficiently localizing specific moments within vast, unstructured video streams remains a fundamental challenge. Video Temporal Grounding (VTG) \cite{Charades,ANet,chatvtg,GroundingGPT,Momentor}, the task of identifying the precise video segment that semantically corresponds to a natural language query, serves as a key step toward bridging this gap. Success in VTG hinges on a model’s ability to comprehend not only static visual content but also intricate temporal dynamics, including action sequences, event duration, and long-range dependencies.

The emergence of Vision-Large Language Models (Vision-LMMs) ~\cite{GPT-4o,Qwen2-VL,Qwen2.5-VL,Qwen3-VL,VTG-LLM} has revolutionized visual understanding ~\cite{videounderstandingsurvey}, exhibiting remarkable zero-shot reasoning ~\cite{chatvtg,vtimecot,videoinsta,zeroshotdensevideocaptioning,MindPalace,VideoTree,TAS,slowfastllavastrongtrainingfreebaseline,TAG,TFVTG,Training-freeVTG} and multi-modal comprehension ~\cite{MVBench,VideoMME,Video-chatgpt,OMG-LLaVA,videounderstandingsurvey}, primarily on static images. However, extending these spatially-oriented architectures to the temporal domain remains non-trivial. As shown in Figure~\ref{fig:intro} (a), Current approaches to incorporate temporal awareness ~\cite{HoPE,KnowingWheretoFocus,Omni-rgpt,Seq2Time} include adding positional encoding (+PE) ~\cite{temporalpositionencodingsurvey,RoPE,V2PE,videorope,VRoPE}, as seen in Qwen2.5-VL ~\cite{Qwen2.5-VL} models, and using text-based timestamps (+TextNum) ~\cite{infinitemotion,egotextvqa,embodied}, as in Qwen3-VL~\cite{Qwen3-VL} models, and visual numbering (+VisualNum) ~\cite{numberit,Vip-llava,dataEfficient3DVisualGrounding}. However, these approaches faces inherent limitations. Positional encoding, though effective for sequence modeling, fail to capture absolute temporal positions essential for grounding specific events and require additional encoding modules. Text-based timestamps (e.g., prompting with ``Frame 1'' or ``1 seconds'') inevitably introduce a rapidly growing amount of textual tokens that leads to increasingly sparse visual attention as the video length increases. 
Visual frame numbering, which overlays timestamps directly on frames, degrades spatial detail and undermines the very visual features that Vision-LMMs rely on for semantic understanding. 

\begin{figure}[t] 
  \centering
  \includegraphics[width=\columnwidth]{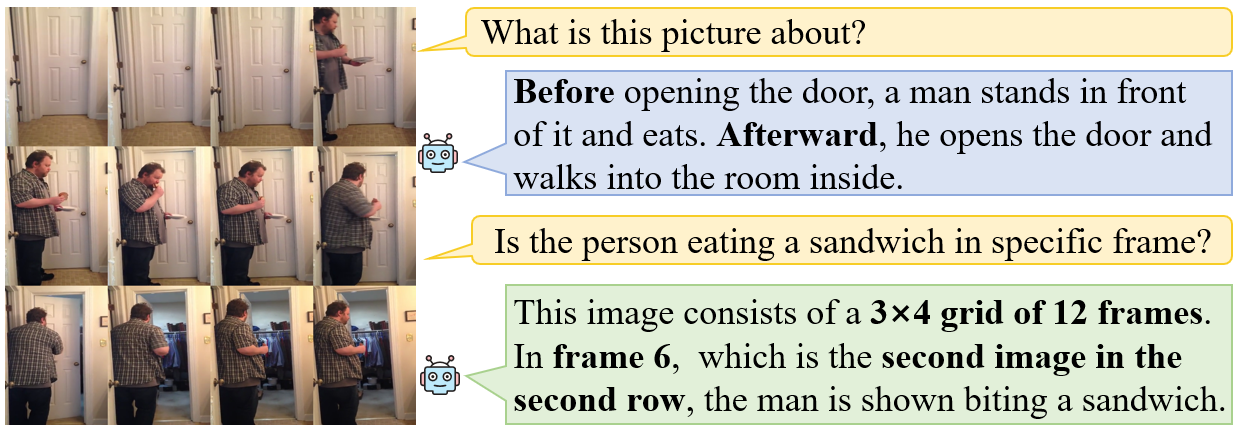}
  \caption{\textbf{Illustration of Qwen2-VL temporal reasoning on 2D grid layouts.} The model correctly infers the temporal order of events (before and after) and accurately identifies the biting action in frame 6.}
  \label{fig:manga}
\end{figure}

To address these challenges, we propose Temporal to Spatial Gridification (T2SGrid), a novel framework that partially reformulates temporal reasoning as a spatial problem. Previously, Vision-LLMs treat a video as a linear sequence of frames. T2SGrid takes a different approach, which partitions the video into configurable temporal windows and arranges the frames within each window into a composite image with grid-structured layout (Figure~\ref{fig:intro} (b)), which we call ``gridification'' of the temporal window. Then, the Vision-LLM processes the grid images rather than the original frames.

The gridification process is central to our approach, establishing a novel paradigm for video input representation. T2SGrid offers several distinct advantages. First, it transforms temporal dynamics within a window into a spatial layout, leveraging spatial attention to enhance the understanding of local temporal dynamics. Second, the row-major frame arrangement in the grid image and the partial overlap between adjacent grid images implicitly function as a form of positional encoding. Crucially, the effectiveness of these advantages is fundamentally enabled by the demonstrated capability of modern Vision-LLMs to interpret grid-based imagery. For instance, as shown in Figure~\ref{fig:manga}, these models can infer relative temporal relationships (e.g., ``before'' and ``after'') by reading the spatial configuration of the grid from top-left to bottom-right, and can also identify the specific order of frames within the grid.




As shown in Figure~\ref{fig:intro} (b), beyond the input gridification, to enhance the model's awareness of global time, we interleave the grid images with text-based timestamps. Unlike existing approaches that assign a timestamp to each individual frame, we associate a single composite text-based timestamp (e.g., ``Frame 0 to 11'') with each grid image. By grouping multiple frames under one temporal descriptor, the model learns to associate a local window of visual content with a unified time interval, further improving temporal understanding. 


In summary, our primary contributions are as follows. (1) We introduce T2SGrid, a novel paradigm that shifts video processing from individual frames to local temporal clips by transforming sequences of frames within a sliding window into a single, composite grid image. (2) Instead of assigning a timestamp to each frame, we use a single composite text timestamp for each grid image, enhancing global temporal awareness. 
(3) Extensive experiments on standard VTG and VQA benchmarks demonstrate that T2SGrid achieves superior performance.

\section{Related Work}
\label{sec:related_work}
\begin{figure*}
  \centering
  \includegraphics[width=\linewidth]{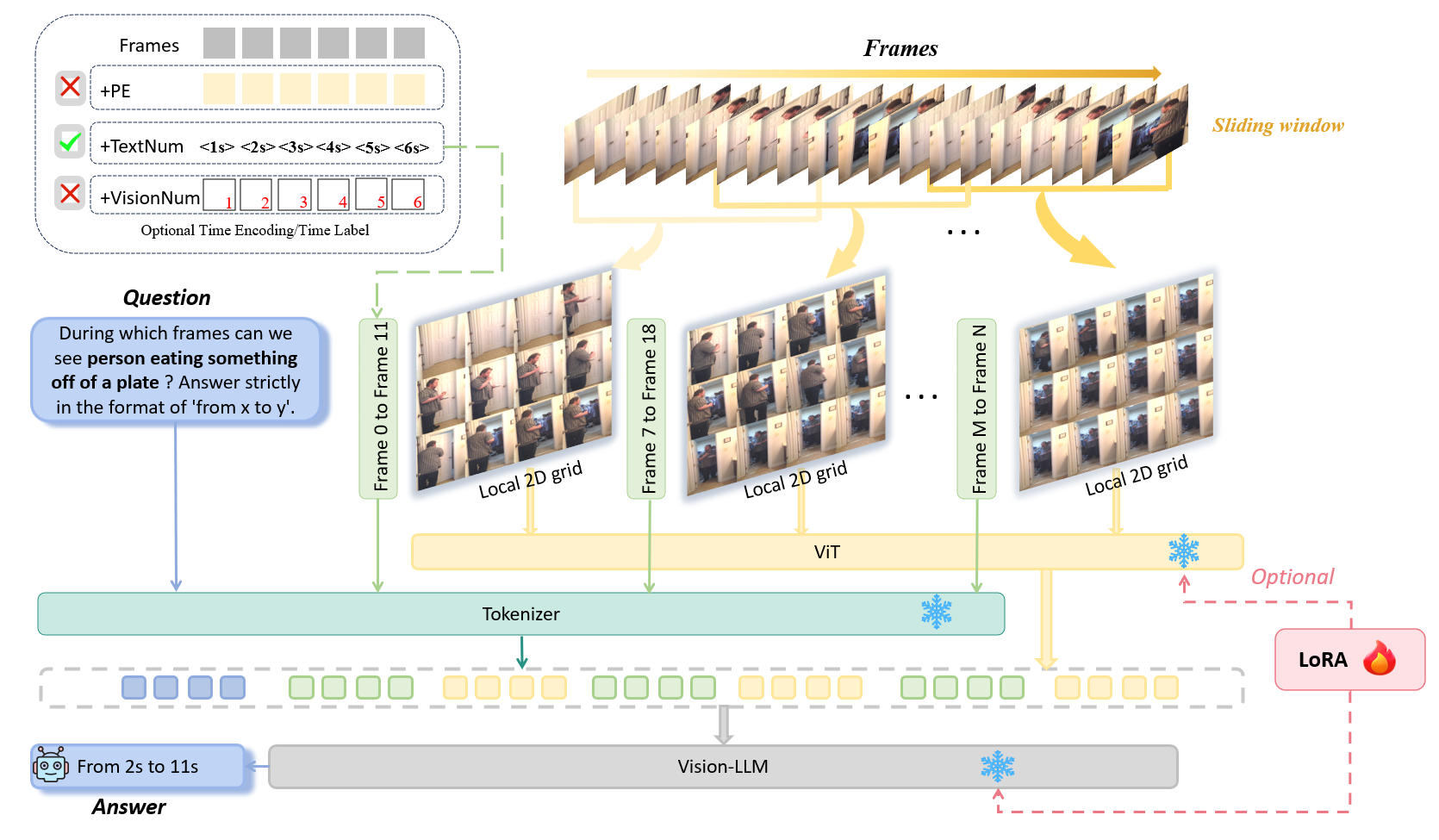}
  \caption{\textbf{Overview of our T2SGrid framework.} The original video frames are first arranged into a 2D grid in a row-major order (\textit{gridication}) to enable spatialized temporal reasoning. A lightweight composite textual timestamp is incorporated to provide global temporal awareness. Our framework can operate in a training-free manner or be further enhanced via LoRA fine-tuning.}
  \label{fig:main}
\end{figure*}
\subsection{Video Temporal Grounding By Vision-LLMs}

Video temporal grounding (VTG) \cite{Temporalsentencegroundingsurvey,UniVTG,End-to-end-VTG,TubeDETR,WhereDoesItExist,interventional,temporallyGrounding,llavamrlargelanguageandvisionassistant} aims to locate the precise time of specific actions or events within a video. For current Vision-LLMs ~\cite{Qwen2-VL,Qwen3-VL,moviechat+,LLaVA-OneVision,Video-chatgpt}, VTG is crucial for both temporal and spatial understanding. Many Vision-LLM-based approaches \cite{videollms,EnrichandDetect,chatvtg,TFVTG,vtggpt,texttoclip,groundedvideollm,GroundingGPT,LITA} have been proposed to tackle the VTG task. Some methods focus on fine-tuning existing models by providing textual timestamps \cite{vtimellmempowerllmgrasp, llavamrlargelanguageandvisionassistant,timer1} and constructing large temporally annotated datasets ~\cite{surveyvideodatasetsgrounded,synopground}. Others design specialized temporal-aware modules using pooling operations \cite{Video-chatgpt, videollama,temporalrelationaware,timesuite,KnowingWheretoFocus} to enhance temporal reasoning. There are also approaches \cite{cpt,ControlMLLM,guidingmedicalvisionlanguagemodels,finegrainedvisualprompting,Video-textprompting} that overlay frame indices ~\cite{Visualpromptingsurvey,setofmark,doesclipknowred,cityllavaefficientfinetuningvlms,groma} onto the original images to provide temporal information and feed the frames sequentially into the model. Although these methods achieve some success, they either require designing task-specific modules for temporal reasoning or compromise spatial information to preserve temporal cues. To address these limitations, we propose T2SGrid, which employs a gridification strategy along with a lightweight textual timestamps to provide global temporal context, eliminating the need for specialized module design or extensive dataset construction, and achieving superior performance.

\subsection{Grid-Based Video Representation}
Although several studies \cite{sharegpt4video,Re-thinking,videopanelslongvideo,Animagegrid,dynimg} have explored merging multiple video frames into a single composite image for video understanding, most of them primarily target video question answering (VQA) ~\cite{vqa,moviechat+,morevqa}. For example, IG-VLM ~\cite{Animagegrid}  selects a small set of keyframes and concatenates them into a single large image to facilitate VQA, yet such a keyframe grid can only model coarse-grained events. DynImg ~\cite{dynimg} enlarges a representative keyframe and appends several smaller frames below it as temporal cues to enhance video comprehension; however, such heuristic designs still fail to preserve fine-grained temporal order, thereby limiting the potential of grid-based representations for temporal reasoning. In contrast, our work is the first to reveal and leverage the intrinsic temporal information encoded within the spatial grid itself, and to apply this property to video temporal grounding, thereby enhancing the generalization capability of gridification for temporal understanding.

\section{Method}
\label{sec:method}
We propose an innovative temporal-to-spatial gridification (T2SGrid) method. In Section~\ref{subsec:attention_analysis}, we first analyze how the gridification mechanism enhances local attention from a spatial perspective and improves temporal perception from the viewpoint of temporal attention. As illustrated in Figure~\ref{fig:main}, our approach consists of two main stages: (1) Sliding-Window Spatiotemporal Gridification, and (2) Temporal Modeling with T2SGrid, which are detailed in Section~\ref{subsec:Gridification} and Section~\ref{subsec:temporal_modeling}, respectively.

\subsection{Attention Analysis}
\label{subsec:attention_analysis}
Current Vision-LLMs process videos by encoding frames into a sequential feature representation. This sequence, concatenated over time, is then aligned with a language query to facilitate understanding. While this approach is effective for content recognition, it presents inherent challenges for tasks requiring fine-grained temporal reasoning. To investigate the underlying limitations of this sequential methodology, we conducted an in-depth analysis of the model's internal attention mechanisms.
Our central hypothesis is that the sequential processing of individual frame features biases Vision-LLMs towards recognizing a series of static spatial configurations rather than understanding the dynamic temporal evolution between them. To test this, we used Qwen2-VL-7B as case studies.

\begin{figure}[t] 
  \centering
  \includegraphics[width=\columnwidth]{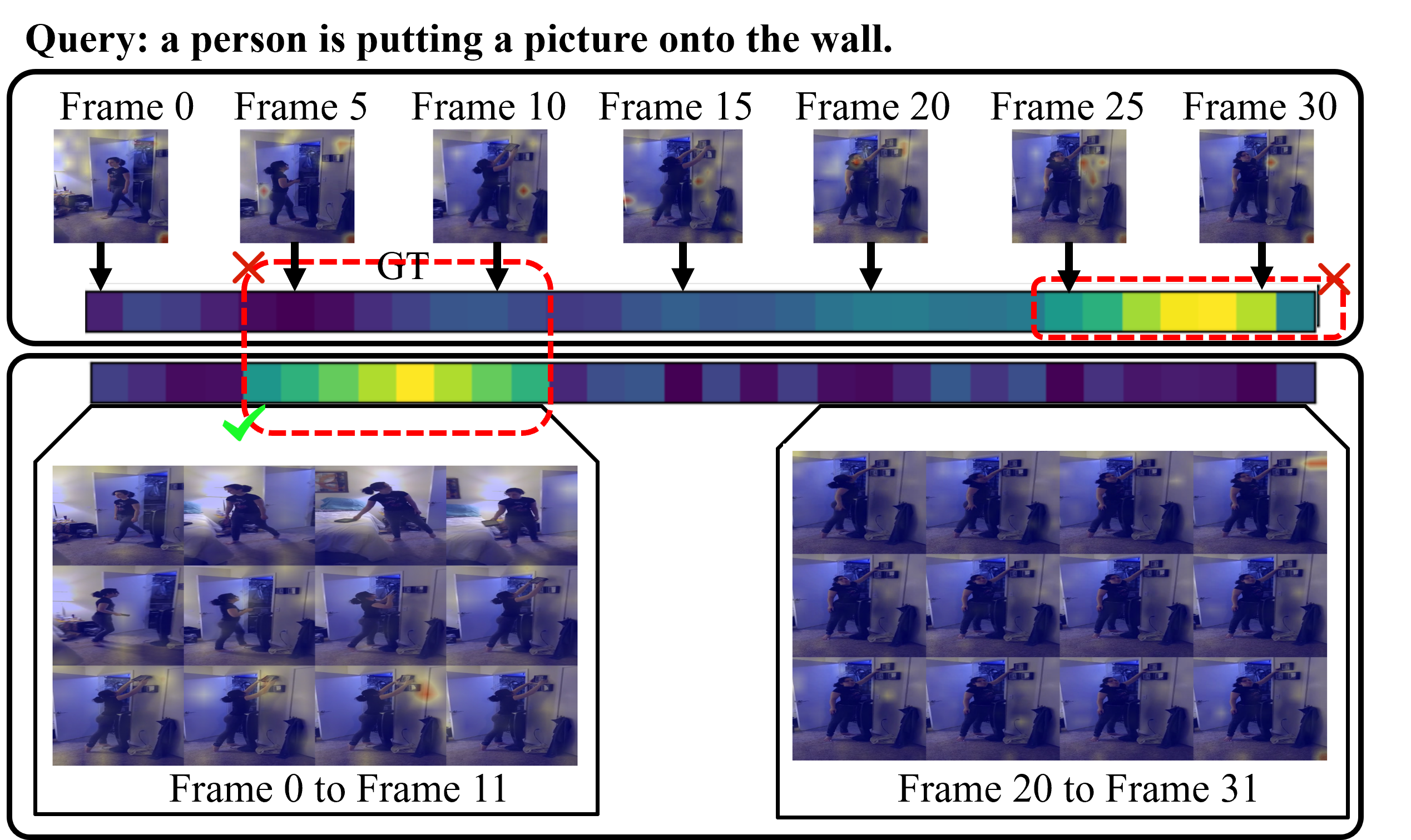}
\caption{\textbf{Top}: spatial and temporal attention with sequential frame input. \textbf{Bottom}: spatial and temporal attention with grid-based input. The grid method captures dynamic actions and maintain focus on the correct temporal intervals.}

  \label{fig:attn}
\end{figure}

We visualize the cross-attention maps between the visual tokens and the query text token, and project them back to the original image or the grid representation. As shown in Figure~\ref{fig:attn}, when given the query “a person is putting a picture onto the wall”, the attention maps under sequential-frame input appear scattered or focus mainly on the person or the picture in a specific frame. This pattern indicates that sequential processing encourages the model to recognize what objects are present, but fails to capture how these objects change over time, resulting in weak sensitivity to motion-induced spatial variations.
In contrast, our T2SGrid representation reorganizes the video into a structured 2D layout that places temporally adjacent patches into spatially coherent neighborhoods. This gridified structure strengthens local spatial attention and enables the model to capture subtle structural changes in the picture frame across neighboring cells. As shown in the Figure~\ref{fig:attn}, the model now attends to how the person moves the picture onto the wall, demonstrating improved temporal understanding through spatial reasoning.
We further average the attention over all visual tokens within each frame to obtain frame-level temporal attention. As shown in the figure, the temporal attention derived from sequential-frame input exhibits peaks that deviate substantially from the Ground Truth (GT), indicating that its temporal modeling is primarily driven by static object saliency rather than the actual progression of the action. In contrast, T2SGrid makes temporal cues more explicit: by embedding temporal adjacency into a 2D spatial layout and combining it with lightweight text-based temporal anchors, the model produces temporally consistent activation patterns that closely align with the GT. Further illustrative examples can be found in the Appendix.

In conclusion, our analyses reveal a fundamental limitation in sequential-frame processing: although the model is proficient at spatial recognition, this ability does not naturally extend to capturing fine-grained temporal progression. Sequential alignment encourages frame-by-frame object matching, causing the model to rely on static object saliency while overlooking subtle inter-frame motion cues. By contrast, the proposed gridification mechanism strengthens local spatial attention, enhances sensitivity to dynamic changes, and produces temporally consistent activation patterns that align closely with the GT. By reframing temporal reasoning into a structured 2D spatial layout, our approach enables the model to leverage its inherent spatial priors to capture genuine temporal cues and achieve more accurate action localization.

\subsection{Sliding Window Spatiotemporal Gridification}
\label{subsec:Gridification}



We first employ a sliding window mechanism to obtain frame sequences, after which we apply spatiotemporal gridification to convert them into a coherent 2D spatial layout.
Concretely, given a video with $T$ frames, we define a temporal window size $k$ and stride $s$. 
The $i$-th window $W_i$ consists of:
\begin{equation}
    W_i = \{ f_{i \times s}, f_{i \times s + 1}, \ldots, f_{i \times s + k - 1} \}.
\end{equation}
For each $W_i$, we spatially rearrange its $k$ \textbf{original resolution frames} into a single composite grid $G_i$. Notably, our method does not compromise spatial resolution; it only performs gridification of the original frames. The layout of this grid is flexible (e.g., $M \times N$) as long as $M \times N = k$. For instance, nine frames can be arranged in a $3 \times 3$ grid. The frames are consistently ordered in a row-major fashion (left-to-right, top-to-bottom).


Temporal localization methods processing input videos that uniformly sample frame sequences are fundamentally equivalent to sliding window processing with stride $s=1$ and window size $k=1$. While capturing per-frame static information, this approach poorly perceives dynamic changes between adjacent frames. Our method overcomes this limitation by setting $k>1$ (e.g., $3\times3$ grid), where each window contains $k$ consecutive frames forming a local temporal segment. When processing center frame $f_t$, the model simultaneously receives its neighborhood ${f_{t-4},\ldots,f_{t+4}}$, establishing complete dynamic context.
To avoid splitting critical actions in short videos, we introduce overlap by setting the sliding-window stride $s < k$, preserving temporal continuity and allowing the model to capture both local dynamics and maintain global consistency. For longer videos, we set $s = k$ to avoid excessive overlap and computational redundancy. This design makes our approach suitable for long videos and robust to a wide range of frame rates (FPS), while maintaining the original video resolution throughout the spatiotemporal gridification process. 

\subsection{Temporal Modeling with T2SGrid}
\label{subsec:temporal_modeling}
\subsubsection{Implicit Temporal Encoding via Grid Layout}

Although the T2SGrid is inherently spatial, arranging frames in a \textbf{row-major order} induces a deterministic mapping that implicitly encodes temporal information. Let $N_c$ denote the number of frames per row. The temporal index $t_f$ of a frame with row-column coordinates $(r_f, c_f)$ is uniquely defined as
\begin{equation}
\label{eq:frame_index}
t_f = r_f \times N_c + c_f,
\end{equation}
reflecting the linear progression along the row-major layout.

The self-attention mechanism, however, operates on patch-level coordinates $(r_p, c_p)$ and their corresponding 2D positional embeddings $E(r_p, c_p)$. A frame-level coordinate can be recovered from patch-level coordinates using
\begin{equation}
\label{eq:patch_to_frame}
r_f = \lfloor r_p / h_{\text{patch}} \rfloor, \quad 
c_f = \lfloor c_p / w_{\text{patch}} \rfloor,
\end{equation}
where $h_{\text{patch}}$ and $w_{\text{patch}}$ are the height and width of a frame in patches. Substituting Equation~\eqref{eq:patch_to_frame} into Equation~\eqref{eq:frame_index} gives
\begin{equation}
\label{eq:patch_index}
t_f = \lfloor r_p / h_{\text{patch}} \rfloor \times N_c + \lfloor c_p / w_{\text{patch}} \rfloor,
\end{equation}
showing that the temporal index is a well-defined function of patch coordinates.

Therefore, the 2D positional embeddings $E(r_p, c_p)$ contain sufficient information for the model to infer the \textbf{temporal order} of frames, enabling implicit temporal reasoning without requiring explicit timestamps or frame identifiers.

Figure~\ref{fig:manga} shows concrete examples of how Qwen2-VL~\cite{Qwen2-VL} interpret the grid structure, providing empirical evidence that the grid layout enables implicit temporal encoding via spatial understanding.


\subsubsection{Absolute Global Temporal Awareness.}
Although the sliding window strategy enhances local temporal reasoning through spatiotemporal grid representations of videos, it loses the absolute temporal position of each segment. This limitation arises from its capture of only relative temporal relationships, such as action continuity between adjacent frames. This limitation prevents the establishment of absolute positioning along the global timeline, as the model loses perception of corresponding absolute time intervals in the original video (e.g., ``From Frame 0 to Frame 8'')  while comprehending dynamic evolution within grid $G_i$. Such deficiency impedes video temporal localization tasks demanding precise timestamp outputs (e.g., ``Xs to Ys'') or global narrative comprehension (e.g., ``What happened at 7 seconds in the video?''), where absolute timing remains essential to satisfy rigid requirements for a temporal reference system.

To preserve global time awareness, we introduce Absolute Time as solution. Before passing each grid image $G_i$ to the LMM, we prepend a textual timestamp to the input prompt:
\begin{equation}
        \text{Prompt}_i = \big[ \text{`` from } T_{\text{start}} \text{ to } T_{\text{end}}.\text{''} \big]; \big[ \text{Image: } G_i \big]
\end{equation}
where $T_{\text{start}}$ and $T_{\text{end}}$ correspond to the start and end time of the $i$-th grid.

For holistic video understanding, multiple grids are organized in an interleaved text-image sequence:
\begin{equation}
\begin{split}
    \big[\text{Text: } T^{(1)}_{\text{start}} \text{ to } T^{(1)}_{\text{end}}\big] &\rightarrow \big[\text{Image: } G_1 \big]; \\
    \big[\text{Text: } T^{(2)}_{\text{start}} \text{ to } T^{(2)}_{\text{end}}\big] &\rightarrow \big[\text{Image: } G_2 \big]; \\
    &\quad\vdots \\
    \big[\text{Text: } T^{(n)}_{\text{start}} \text{ to } T^{(n)}_{\text{end}}\big] &\rightarrow \big[\text{Image: } G_n \big].
\end{split}
\end{equation}

These absolute timestamps form a continuous temporal chain across the video, enabling the model to establish sequential relationships while preserving temporal coherence. By interleaving grids with global textual time  and leveraging cross-attention mechanisms, the model can reason not only about the dynamics within each grid but also about their placement along the full video timeline.
\section{Experiment}
\label{sec:experiment}
\begin{table*}[t]
\renewcommand\arraystretch{0.95}
\centering
\caption{\textbf{Performance comparison on the video temporal grounding task with prior state-of-the-art methods.} \textit{T2SGrid} denotes the use of spatiotemporal gridification with global time awareness, while \textit{T2SGrid-FT} indicates fine-tuning with the instruction dataset augmented using our T2SGrid method.}
\resizebox{\textwidth}{!}{
\begin{tabular}{l|cccc|cccc}
\toprule
\multirow{2}{*}{Model} & \multicolumn{4}{c|}{Charades-STA} & \multicolumn{4}{c|}{ActivityNet}\\
                 & R@0.3 & R@0.5 & R@0.7 & mIoU & R@0.3 & R@0.5 & R@0.7 & mIoU \\ 
\midrule
\multicolumn{9}{c}{\textit{VTG-Tuned Vision-LLMs}} \\
\midrule
GroundingGPT~\cite{GroundingGPT} & - & 29.6 & 11.9 & - & - & - & - & - \\
LITA~\cite{LITA} & - & - & - & - & - & 25.9 & - & 28.6  \\
VTG-LLM~\cite{VTG-LLM} & 52.0 & 33.8 & 15.7 & - & - & - & - & -  \\
TimeChat~\cite{TimeChat} & 47.7 & 22.9 & 12.5 & 30.6 & 30.2 & 16.9 & 8.2 & 21.8 \\
VTimeLLM~\cite{vtimellmempowerllmgrasp} & 51.0 & 27.5 & 11.4 & 31.2 & 44.0 & 27.8 & 14.3 & 30.4  \\
Momentor~\cite{Momentor} & 42.9 & 23.0 & 12.4 & 29.3 & 42.6 & 26.6 & 11.6 & 28.5  \\
HawkEye~\cite{Hawkeye} & 50.6 & 31.4 & 14.5 & 33.7 & 49.1 & 29.3 & 10.7 & 32.7 \\
TimeSuite~\cite{timesuite} & \underline{69.9} & \underline{48.7} & \underline{24.0} & - & - & - & - & -  \\
TRACE~\cite{guo2024trace} & - & 40.3 & 19.4 & - & - & \underline{37.7} & \underline{24.0} & \underline{39.0}  \\
NumPro~\cite{numberit} & 63.8 & 42.0 & 20.6 & \underline{41.4} & \underline{55.6} & 37.5 & 20.6 & 38.8  \\
\midrule
\multicolumn{9}{c}{\textit{General Vision-LLMs}} \\
\midrule
GPT-4o~\cite{GPT-4o} & 55.0 & 32.0 & 11.5 & 35.4 & 33.3 & 21.2 & 10.4 & 23.7 \\
\rowcolor{gray!20} ~~~~~~\textit{+T2SGrid} & $57.3_\impro{2.3}$ & $36.7_\impro{4.7}$ & $14.8_\impro{3.3}$ & $36.9_\impro{1.5}$ & $47.5_\impro{14.2}$ & $32.1_\impro{10.9}$ & $19.7_\impro{9.3}$ & $35.6_\impro{11.9}$ \\
\midrule
Qwen3-VL-8B~\cite{Qwen3-VL} & 69.3 & 43.4 & 17.5 & 43.1 & 39.0 & 26.2 & 15.8 & 29.3 \\
\rowcolor{gray!20} ~~~~~~\textit{+T2SGrid} & $71.4_\impro{2.1}$ & $47.0_\impro{3.6}$ & $20.7_\impro{3.2}$ & $44.9_\impro{1.8}$ & $44.8_\impro{5.8}$ & $26.8_\impro{0.6}$ & $16.1_\impro{0.3}$ & $32.5_\impro{3.2}$ \\
\midrule
LLaVA-OneVision-1.5-8B~\cite{LLaVA-OneVision} & 19.8 & 6.7 & 2.3 & 14.5 & 9.2 & 4.9 & 2.1 & 7.5 \\
\rowcolor{gray!20} ~~~~~~\textit{+T2SGrid} & $45.0_\impro{25.2}$ & $26.3_\impro{19.6}$ & $11.9_\impro{9.6}$ & $28.8_\impro{14.3}$  & $31.5_\impro{22.3}$ & $17.4_\impro{12.5}$ & $10.4_\impro{8.3}$ & $23.6_\impro{16.1}$ \\
\midrule

Qwen2-VL-7B~\cite{Qwen2-VL} & 8.7 & 5.4 & 2.4 & 7.9 & 17.0 & 9.4 & 3.9 & 12.5 \\
\rowcolor{gray!20} ~~~~~~\textit{+T2SGrid} & $70.1_\impro{61.4}$ & $46.7_\impro{41.3}$ & $20.1_\impro{17.7}$ & $44.3_\impro{36.4}$ & $46.2_\impro{29.2}$ & $27.2_\impro{17.8}$ & $15.4_\impro{11.5}$ & $33.3_\impro{20.8}$ \\
\rowcolor{gray!20} ~~~~~~\textit{+T2SGrid-FT} & $\textbf{76.9}_\impro{68.2}$ & $\textbf{60.6}_\impro{55.2}$ & $\textbf{35.9}_\impro{33.5}$ & $\textbf{53.2}_\impro{45.3}$ & $\textbf{64.4}_\impro{47.4}$ & $\textbf{48.4}_\impro{39.0}$ & $\textbf{29.5}_\impro{25.6}$ & $\textbf{46.7}_\impro{34.2}$  \\
\bottomrule
\end{tabular}}
\vspace{-10pt}
\label{table:main}
\end{table*}

We evaluate our model on standard Video Temporal Grounding benchmarks. Following prior work~\cite{vtimellmempowerllmgrasp,TimeChat,Momentor,Hawkeye}, we evaluate our method on the test sets of Charades-STA~\cite{Charades} and ActivityNet~\cite{ANet}. The evaluation metrics include mean Intersection over Union (mIoU) and Recall@1 at multiple IoU thresholds (R@\(m\)), where \(m \in \{0.3, 0.5, 0.7\}\), consistent with previous studies ~\cite{TimeChat,vtimellmempowerllmgrasp}. 


\subsection{Experiment setup}
\paragraph{Training dataset and benchmark}  
We adopt a mixed training set, combining the training splits of ActivityNet-Captions~\cite{ANet} and Charades~\cite{Charades}, containing 18K videos and 50K question–answer pairs. Each video in our dataset is augmented and arranged using our T2SGrid method.
The question–answer pairs follow a consistent template: questions are formatted as ``During which frames can we see {query}?'', and answers are formatted as ``From x to y'', where x and y denote the start and end frame numbers of the queried event.

\vspace{-15pt}
\paragraph{Training Detail}
We use Qwen2-VL-7B as the base model, which lacks temporal encoding and therefore enables a fair comparison of temporal strategies. The model is trained for 3 epochs with a batch size of 32 and a learning rate of $2\times10^{-5}$. We adopt LoRA fine-tuning (rank 64, $\alpha=128$) on all linear layers. All experiments are conducted on 4$\times$A100 GPUs. Charades-STA is trained and evaluated under the overlap configuration (g43\_s7), whereas ActivityNet-Captions uses the no-overlap configuration (g43\_s12). Additional details are provided in the Appendix.
\subsection{Main result}

\subsubsection{Superior performance on VTG datsets}
Table~\ref{table:main} reports the performance of our T2SGrid and T2SGrid-FT methods compared with previous approaches on video temporal grounding benchmarks.
After integrating the proposed T2SGrid temporal encoding, both open-source and closed-source models exhibit consistent improvements. 

Specifically, GPT-4o, which already possesses strong temporal reasoning capability, achieves further gains with T2SGrid, showing that our method can enhance well-trained multimodal LLMs.
Qwen3-VL-8B, a state-of-the-art vision-LLM utilizing textual timestamps for temporal modeling, also benefits modestly; for instance, its Charades-STA R@0.3 increases from 69.3 to 71.4. The smaller gain mainly arises from a slight conflict between our local temporal encoding and its original timestamp-based scheme.
In contrast, Qwen2-VL-7B, which lacks explicit temporal encoding, exhibits a significant performance boost when combined with T2SGrid, reaching 70.1 R@0.3 and 44.3 mIoU on Charades-STA thatsurpassing several VTG-tuned video LLMs.

Finally, LLaVA-OneVision1.5-8B, a model trained solely on static images, achieves remarkable improvements with T2SGrid: absolute gains of 25.2, 19.6, 9.6, and 14.3 on R@0.3, R@0.5, R@0.7, and mIoU, respectively.
On ActivityNet, the corresponding improvements are 22.3, 12.5, 8.3, and 16.1.
These results confirm that T2SGrid effectively leverages the spatial reasoning capability of image-based models to enable temporal understanding.

To further validate the effectiveness of our approach, We fine-tune Qwen2-VL-7B on a dataset augmented and arranged using our T2SGrid framework, resulting in the best mIoU scores on Charades-STA and ActivityNet, reaching 53.2 and 46.7, respectively.

\begin{table*}[t]
\centering
\caption{\textbf{Training effectiveness of our method.} T2SGrid-FT consistently outperforms standard fine-tuning (FT) when trained on the same data with identical training settings, demonstrating the benefits of our approach.}

\resizebox{\textwidth}{!}{
\begin{tabular}{l|llll|llll}
\toprule
\multirow{2}{*}{Model} & \multicolumn{4}{c|}{Charades-STA} & \multicolumn{4}{c}{ActivityNet} \\
& R@0.3 & R@0.5 & R@0.7 & mIoU & R@0.3 & R@0.5 & R@0.7 & mIoU \\
\midrule
LLaVA-OneVision-1.5-8B~\cite{LLaVA-OneVision} & 19.8 & 6.7 & 2.3 & 14.5 & 9.2 & 4.9 & 2.1 & 7.5 \\

~~~~~~\textit{+FT} & 70.6 & 52.9 & 29.8 & 48.4 & 50.1 & 35.1 & 18.5 & 33.4 \\
\rowcolor{gray!20} ~~~~~~\textit{+T2SGrid-FT} & $73.9_\impro{3.3}$ & $57.7_\impro{4.8}$ & $32.7_\impro{2.9}$ & $50.8_\impro{2.4}$ & $63.5_\impro{13.4}$ & $45.4_\impro{10.3}$ & $26.2_\impro{7.7}$ & $44.6_\impro{10.2}$ \\

\midrule
Qwen2-VL-7B~\cite{Qwen2-VL} & 8.7 & 5.4 & 2.4 & 7.9 & 17.0 & 9.4 & 3.9 & 12.5 \\
~~~~~~\textit{+FT} & 73.1 & 56.8 & 32.7 & 50.4 & 54.9 & 37.2 & 19.2 & 37.7 \\
\rowcolor{gray!20} ~~~~~~\textit{+T2SGrid-FT} & $76.8_\impro{3.7}$ & $60.6_\impro{3.8}$ & $35.9_\impro{3.2}$ & $53.2_\impro{2.8}$ & $64.4_\impro{9.5}$ & $48.4_\impro{9.8}$ & $29.5_\impro{10.3}$ & $46.7_\impro{9.0}$ \\
\bottomrule
\end{tabular}
}
\vspace{-10pt}
\label{table:ablation_ft}
\end{table*}
\begin{table*}[h]
\centering
\caption{\textbf{Performance comparison on Video-MME, MVBench, and VideoInstruct benchmarks}. Our T2SGrid method consistently improves temporal perception and action understanding, demonstrating strong generalization QA tasks.}

\label{tab:benchmark_results}
\resizebox{\textwidth}{!}{
\begin{tabular}{lcccccccc cc}
\toprule
\multirow{3}{*}{Model} 
 & \multicolumn{3}{c}{Video MME} 
 & \multicolumn{5}{c}{MVbench}
 & \multicolumn{2}{c}{VideoInstruct} \\

\cmidrule(lr){2-4}
\cmidrule(lr){5-9}
\cmidrule(lr){10-11}

 & \begin{tabular}[c]{@{}c@{}}Temporal\\Perception\end{tabular}
 & \begin{tabular}[c]{@{}c@{}}Temporal\\Reasoning\end{tabular}
 & All
 & \begin{tabular}[c]{@{}c@{}}Action\\Sequence\end{tabular}
 & \begin{tabular}[c]{@{}c@{}}Fine-grained\\Action\end{tabular}
 & \begin{tabular}[c]{@{}c@{}}State\\Change\end{tabular}
 & \begin{tabular}[c]{@{}c@{}}Scene\\Transition\end{tabular}
 & All
 & \begin{tabular}[c]{@{}c@{}}Temporal\\Understanding\end{tabular}
 & \begin{tabular}[c]{@{}c@{}}Detail\\Orientation\end{tabular} \\

\midrule
Qwen2-VL-7B 
 & 60.0 & 41.7 & 63.3 
 & 60.5 & 46.5 & 44.0 & 79.5 & 51.7 
 & 2.47 & 2.57 \\

+T2SGrid       
& \textbf{74.5} & \textbf{50.2} & \textbf{64.1} 
& \textbf{71.5} & \textbf{49.5} & \textbf{51.5} & \textbf{89.0} & \textbf{58.3}
& \textbf{2.52} & \textbf{2.69} \\
\bottomrule
\end{tabular}
}
\end{table*}

\subsubsection{Training Effectiveness}
In Table \ref{table:ablation_ft}, we observe consistent performance gains when incorporating T2SGrid during fine-tuning (+T2SGrid-FT). Both the standard fine-tuning (FT) and T2SGrid-FT experiments are conducted on the same training data, ensuring a fair comparison.
For LLaVA-OneVision-1.5-8B, T2SGrid yields clear improvements of +4.8 R@0.5 and +2.4 mIoU on Charades-STA, and large boosts of +13.4 R@0.3 and +10.2 mIoU on ActivityNet.
For Qwen2-VL-7B, it further improves results by +3.8 R@0.5 and +2.8 mIoU on Charades-STA, and by +9.8 R@0.5, +10.3 R@0.7, and +9.0 mIoU on ActivityNet.
Overall, models fine-tuned with our T2SGrid framework not only achieve consistently higher scores across all metrics and datasets but also demonstrate stronger temporal grounding capability, confirming the superior training effectiveness of our approach.

\subsubsection{Qualitative Results}
\begin{figure*}
  \centering
  \includegraphics[width=\linewidth]{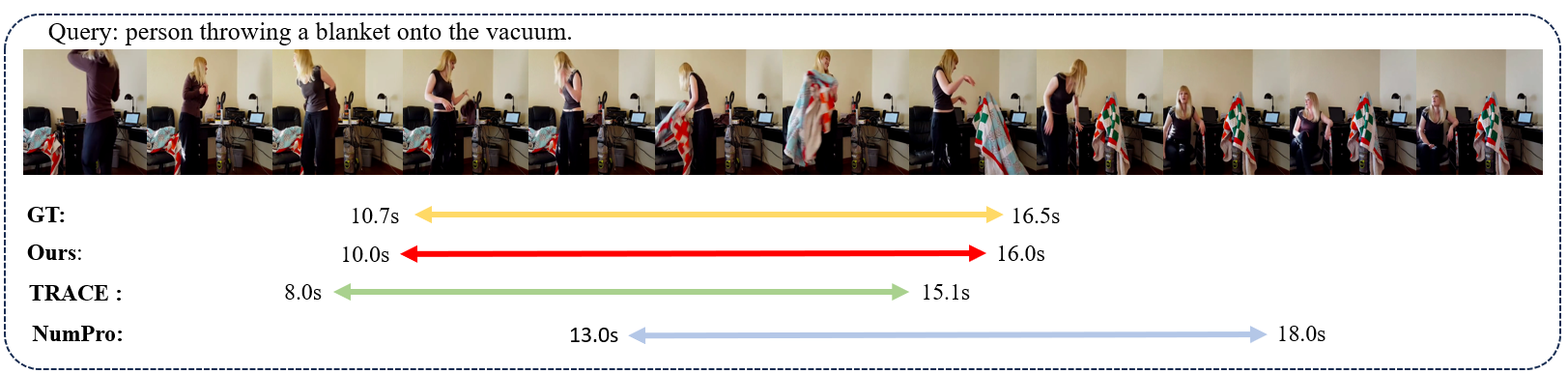}
  \caption{Qualitative Comparison with State-of-the-Art. Our method outperforms TRACE ~\cite{guo2024trace} and Numpro~\cite{numberit} on Charades by accurately identifying event boundaries in challenging scenes.}
  \label{fig:visual}
\end{figure*}

Figure~\ref{fig:visual} illustrates a visualization on the Charades dataset, comparing our method with previous approaches. Given the query “person throwing a blanket onto the vacuum”, other methods predict incorrect temporal intervals, whereas our method accurately captures the start and end times of the action, demonstrating the effectiveness of T2SGrid. Additional examples are provided in the Appendix.

\subsubsection{Effectiveness on Long-Video and VQA Tasks}
To further evaluate the effectiveness of our T2SGrid method in enhancing temporal understanding and action reasoning, we conduct experiments on three video understanding benchmarks: VideoMME~\cite{VideoMME}, MVBench~\cite{MVBench}, and VideoInstruct~\cite{Video-chatgpt}. Among them, VideoMME is a long-form video QA benchmark that further demonstrates our method’s ability to reason over long-duration videos.

As shown in Table~\ref{tab:benchmark_results}, the T2SGrid-enhanced model consistently outperforms the base Qwen2-VL-7B model across all metrics. On VideoMME, it improves temporal perception and temporal reasoning by 14.5 and 8.5, respectively. Similarly, on MVBench, TGrid yields consistent gains across all subcategories including action sequence, fine-grained action, state change, and scene transition, raising the overall score from 51.7 to 58.3. On the VideoInstruct dataset, our method also achieves higher scores in both temporal understanding and detail orientation, confirming that T2SGrid effectively enhances temporal reasoning and fine-grained perception across diverse video QA tasks. 

These results demonstrate the generality and robustness of T2SGrid across both long and short video QA tasks, highlighting its strong ability to improve temporal perception and question answering performance.

\subsection{Ablation studies}
To isolate the contributions of each component of our method, we conduct a series of ablation studies on the Charades dataset using Qwen2-VL-7B.

\paragraph{Component Ablation.} 
\begin{table}[h]
\vspace{-10pt}
\centering
\caption{\textbf{Ablation study of key components:} composite textual timestamps (ComTextNum), sliding window, and grid layout. Each component contributes to improved temporal modeling and overall grounding performance.}

\label{tab:abla_component}
\resizebox{\linewidth}{!}{
\begin{tabular}{ccc cccc}
\toprule
\textbf{ComTextNum} & \textbf{Sliding Window} & \textbf{Grid} & \textbf{R1@0.3} & \textbf{R1@0.5} & \textbf{R1@0.7} & \textbf{mIoU} \\
\midrule
\xmark & \xmark & \xmark & 8.7 & 5.4 & 2.4 & 7.9 \\
\checkmark & \xmark & \xmark & 53.5 & 23.2 & 7.9 & 32.9 \\
\checkmark & \checkmark & \xmark & \underline{58.3} & \underline{35.1} & \underline{13.6} & \underline{36.5} \\
\checkmark & \checkmark & \checkmark & \textbf{70.1} & \textbf{46.7} & \textbf{20.1} & \textbf{44.3} \\
\bottomrule
\end{tabular}
}
\end{table}
We conduct a comprehensive ablation study to evaluate the contribution of each component in our temporal encoding framework. ComTextNum denotes the use of a textual timestamp inserted before each grid element, providing an explicit temporal anchor for the model. Sliding Window refers to partitioning the sequential frames into temporally localized windows, while Grid represents the transformation of frames within each window into a 2D spatial grid.

As shown in Table~\ref{tab:abla_component}, introducing ComTextNum substantially improves temporal information, boosting mIoU from 7.9 to 32.9. Incorporating the sliding window mechanism further enhances local spatiotemporal attention, raising mIoU to 36.5. When we additionally reformulate each temporal window into a 2D grid structure, the model achieves the largest performance gain, reaching 70.1 R1@0.3 and 44.3 mIoU. These improvements demonstrate that local implicit temporal encoding within the grid provides advantages over TextNum, validating the effectiveness of our core grid-based approach.

\paragraph{Temporal Modeling Method Ablation.}

\begin{table}[t]
\centering
\caption{Comparison of different temporal modeling strategies applied to the Qwen2-VL-7B baseline, including position encoding (PE), textual timestamping (TextNum), visual frame-level numeric labeling (VisualNum), and our proposed T2SGrid framework. mToken and mTime/s denote the average token count and inference time per sample, respectively.
}
\label{tab:time_encoding_ablation}
\resizebox{0.48\textwidth}{!}{
\begin{tabular}{l | c c c c | c c}
\toprule
\textbf{Temporal Modeling} & \textbf{R1@0.3} & \textbf{R1@0.5} & \textbf{R1@0.7} & \textbf{mIoU} 
& \textbf{mToken} & \textbf{mTime/s} \\
\midrule
PE  & 53.1 & 28.4 & 10.6 & 33.8 & 5760.4 & 1.69\\
TextNum    & 53.5 & 23.2 & 7.9 & 32.5 & 5791.2 & 1.45\\
VisualNum  & 60.7 & 36.8 & 15.9 & 38.5 & 5760.4 & 2.17\\
Ours (wo/ overlap)  & \underline{64.5} & \underline{42.9} & \underline{18.1} & \underline{41.2} & 5766.1 & 1.43 \\
Ours (w/ overlap)  & \textbf{70.2} & \textbf{46.7} & \textbf{20.1} & \textbf{44.3} & 7909.7 & 2.31\\
\bottomrule
\end{tabular}
}
\vspace{-15pt}
\end{table}
We conduct an ablation study to evaluate four temporal encoding strategies on the Qwen2-VL-7B baseline: PE, TextNum, VisualNum, and Ours. PE encodes temporal order through positional encoding; TextNum adds textual timestamps before each frame or grid element; VisualNum draws numeric indices directly on frames. As shown in Table~\ref{tab:time_encoding_ablation}, Under the non-overlap setting, our method delivers the best overall performance while reducing inference time by 34.1\% over VisualNum. With overlap, it trades only 6\% additional time for a 14\% performance gain. 

\vspace{-10pt}
\paragraph{Ablation Study on Grid Configuration}

We conduct ablations over different grid configurations, denoted as \texttt{g\{col\}\{row\}\_s\{stride\}} where $k=\text{col}\times\text{row}$. 

In Stage~1, we vary both grid size and stride (with $k=s$). As shown in Table~\ref{tab:grid_ablation}, performance improves as the window size increases: R1@0.3 rises from 53.5 (g11\_s1) to 63.7 (g33\_s9), and mIoU increases from 32.9 to 41.2. The best configuration in this stage is g43\_s12, reaching 64.5 R1@0.3 and 41.2 mIoU. However, excessively large grids (e.g., g44\_s16) lead to degraded performance, with mIoU dropping to 35.9.
Similarly, g43 is also the optimal configuration on ActivityNet.

Notably, under the no-overlap setting, our method does not reduce the resolution of the original frames, with the average token count (mToken) essentially the same as sequential frame input (g11\_s1), since we simply concatenate full resolution frames into a single grid. This design already yields strong gains.

\begin{table}[h]
\centering
\caption{\textbf{Ablation results of grid configurations on Qwen2-VL-7B.} \texttt{g\{col\}\{row\}\_s\{stride\}} denoting grid size and stride. Stage~1 varies size and stride, and Stage~2 refines stride for g43.}
\resizebox{0.48\textwidth}{!}{
\begin{tabular}{lcccc|cc}
\toprule
\textbf{Config} & \textbf{R1@0.3} & \textbf{R1@0.5} & \textbf{R1@0.7} & \textbf{mIoU} & \textbf{mToken} & \textbf{mTime/s} \\
\midrule
\multicolumn{7}{c}{Stage 1: Exploring Grid size and stride} \\
g11\_s1 & 53.5 & 23.2 & 7.9 & 32.9 & 5791.2 & 1.45 \\
g22\_s4 & 56.5 & 26.9 & 8.7 & 34.2 & 5789.0 & 1.43\\
g23\_s6 & 61.6 & 38.7 & 16.2 & 38.8 & 5784.3 & 1.42\\
g32\_s6 & 61.8 & 38.8 & 16.9 & 39.0 & 5784.3 & 1.43\\
g33\_s9 & 63.7 & \underline{42.8} & \textbf{18.4} & \underline{41.1} & 5782.1 & 1.43\\
g34\_s12 & 64.1 & 42.7 & 18.0 & \underline{41.1} & 5781.4 & 1.42\\
g43\_s12 & \textbf{64.5} & \textbf{42.9} & \underline{18.1} & \textbf{41.2} & 5781.4 & 1.42\\
g44\_s16 & 61.6 & 29.1 & 10.4 & 35.9 & 5779.7 & 1.41\\
\midrule
\multicolumn{7}{c}{Stage 2: Fine-tuning stride for optimal Grid size (g43)} \\
g43\_s5 & 68.9 & 45.1 & 18.2 & 43.2 & 8971.4 & 2.39\\
g43\_s6 & \underline{69.5} & \underline{46.4} & \underline{19.6} & \underline{43.7} & 8357.2 & 2.34\\
g43\_s7 & \textbf{70.2} & \textbf{46.7} & \textbf{20.1} & \textbf{44.3} & 7909.6 & 2.31\\
g43\_s12 & 64.5 & 42.9 & 18.1 & 41.2 & 5781.4 & 1.42\\
\bottomrule
\end{tabular}
}
\label{tab:grid_ablation}
\vspace{-15pt}
\end{table}

In Stage~2, we fix the grid size to g43 and adjust the stride to introduce overlap. Overlapping windows yield consistent improvements: as the stride decreases from 12 to 7, R1@0.3 increases from 64.5 to 70.2, and mIoU improves from 41.2 to 44.3. The best configuration is g43\_s7, which increases the number of tokens modestly but brings best performance.

\vspace{-5pt}
\section{Conlcusion}
\label{sec:conlcusion}
In this work, we introduced T2SGrid, a novel framework designed to tackle the challenge of video temporal grounding (VTG) for Vision-LLMs. Our core idea is to reformulate temporal understanding as a spatial reasoning problem. By ``gridifying'' frames from a sliding temporal window into a single composite 2D image, T2SGrid directly leverages the powerful, pre-existing spatial attention mechanisms of 2D Vision Transformers. This approach enhances the model’s ability to capture fine-grained local temporal dynamics without the need for task-specific temporal modules. We further combined this local gridification with a global absolute time modeling strategy, using composite text timestamps for each grid. Together, these components provide a stronger temporal encoding mechanism. Extensive experiments on standard VTG and VQA benchmarks demonstrate that T2SGrid achieves superior performance.
\section*{Acknowledgments}
This research was supported by Guangdong Laboratory of Artificial Intelligence and Digital Economy (SZ)
{
    \small
    \bibliographystyle{ieeenat_fullname}
    \bibliography{main}
}


\end{document}